%% file: camera_ready_v2.tex
\documentclass[10pt,twocolumn,letterpaper]{article}

\usepackage{iccv}
\usepackage{times}
\usepackage{epsfig}
\usepackage{graphicx}
\usepackage{amsmath}
\usepackage{amssymb}
\usepackage{enumitem}
\usepackage{algorithm}
\usepackage{algpseudocode}
\usepackage[]{multirow}
\usepackage[nocompress]{cite}
\usepackage[font=footnotesize]{caption}
\usepackage{booktabs}
\usepackage[accsupp]{axessibility}
\makeatletter
\@namedef{ver@everyshi.sty}{}
\makeatother
\usepackage{pgfplots}
\usetikzlibrary{patterns}
\usepackage{standalone}
\makeatletter
\renewcommand{\Function}[2]{%
  \csname ALG@cmd@\ALG@L @Function\endcsname{#1}{#2}%
  \def\jayden@currentfunction{#1}%
}
\newcommand{\funclabel}[1]{%
  \@bsphack
  \protected@write\@auxout{}{%
    \string\newlabel{#1}{{\jayden@currentfunction}{\thepage}}%
  }%
  \@esphack
}
\DeclareMathOperator*{\argmax}{argmax}
\usepackage{hyperref}
\hypersetup{pagebackref=true,breaklinks=true,letterpaper=true,colorlinks,bookmarks=false}

\iccvfinalcopy 

\def\iccvPaperID{10357} 

\ificcvfinal\fi

\begin{document}

\title{Attention-based Multi-Reference Learning for Image Super-Resolution}

\author{Marco Pesavento \hspace{3cm} Marco Volino \hspace{3cm} Adrian Hilton \\ 
Centre for Vision, Speech and Signal Processing\\
University of Surrey, UK\\
{\tt\small \{m.pesavento,m.volino,a.hilton\}@surrey.ac.uk}


 }

\maketitle
\ificcvfinal\thispagestyle{empty}\fi

\begin{abstract}
This paper proposes a novel Attention-based Multi-Reference Super-resolution network (AMRSR) that, given a low-resolution image, learns to adaptively transfer the most similar texture from multiple reference images to the super-resolution output whilst maintaining spatial coherence. 
The use of multiple reference images together with attention-based sampling is demonstrated to achieve significantly improved performance over state-of-the-art reference super-resolution approaches on multiple benchmark datasets.
Reference super-resolution approaches have recently been proposed to overcome the ill-posed problem of image super-resolution by providing additional information from a high-resolution reference image. 
Multi-reference super-resolution extends this approach by providing a more diverse pool of image features to overcome the inherent information deficit whilst maintaining memory efficiency. 
A novel hierarchical attention-based sampling approach is introduced to learn the similarity between low-resolution image features and multiple reference images based on a perceptual loss. 
Ablation demonstrates the contribution of both multi-reference and hierarchical attention-based sampling to overall performance. 
Perceptual and quantitative ground-truth evaluation demonstrates significant improvement in performance even when the reference images deviate significantly from the target image. The project
website can be found at \href{https://marcopesavento.github.io/AMRSR/}{https://marcopesavento.github.io/AMRSR/}
\end{abstract}
\section{Introduction}
\noindent
Image super-resolution (SR) aims to estimate a perceptually plausible high-resolution (HR) image from a low-resolution (LR) input image ~\cite{yang2014single}. This problem is ill-posed due to the inherent information deficit between LR and HR images. Classic super-resolution image processing~\cite{pandey2018compendious} and deep learning based approaches~\cite{wang2020deep} result in visual artefacts for large up-scaling factors ($4\times$).
To overcome this limitation, recent research has introduced the sub-problem of reference image super-resolution (RefSR)~\cite{boominathan2014improving,yue2013landmark,zheng2018crossnet}. Given an input LR image and a similar HR reference image, RefSR approaches estimate a SR image. Reference super-resolution with a single reference image has been demonstrated to improve performances over general SR methods achieving large up-scaling with reduced visual artefacts. 
\begin{table}[t]
\centering
\resizebox{1.\linewidth}{!}{\input{amrsr_arxive/image_files/intro2}}
 \vspace{-1.2em}
 
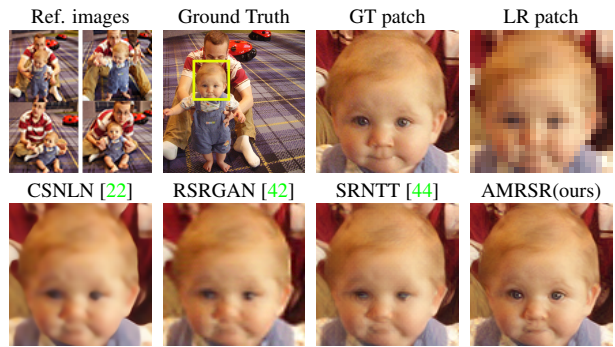
\captionof{figure}{The proposed network exploits $N_M=4$ reference images (top left) to super-resolve the LR input (top right). Its SR output has the best visual quality compared to other state-of-the-art methods.}
 \vspace{-1.7em}
 \label{fig:intro}
\end{table}
\\In this paper we generalise reference super-resolution to use multiple reference images giving a pool of image features and propose a novel attention-based sampling approach to learn the perceptual similarity between reference features and the LR input. 
%
The proposed attention-based multiple-reference super-resolution network (AMRSR) is designed to allow multiple HR reference images by introducing a hierarchical attention-based mapping of LR input feature subvectors into HR reference feature vectors, focusing the learning attention on the LR input. This allows training with multiple HR reference images which would not be possible with a naive extension of existing single-reference super-resolution methods without a significant increase in memory footprint.
%
Figure~\ref{fig:intro} qualitatively illustrates the performance of the proposed AMRSR approach against state-of-the-art single-image super-resolution (CSNLN~\cite{mei2020image}, RSRGAN~\cite{zhang2019ranksrgan}) and RefSR (SRNTT~\cite{ZhangSRNTT2019}) approaches. Given $N_M$ reference images, AMRSR produces a $4\times$ SR image which is perceptually plausible and has a similar level of detail to the ground-truth HR image.   
The primary contributions of the AMRSR approach presented in this paper are:
\begin{itemize}[noitemsep]
    \item Generalisation of single reference super-resolution to multiple reference images whilst improving memory efficiency thanks to a part-based mechanism.
    \item Hierarchical attention-based adaptive sampling for perceptual similarity learning between low-resolution image features and multiple HR reference images.
    \item Improved quantitative and perceptual performance for image super-resolution compared with state-of-the-art single-image RefSR. 
\end{itemize}
AMRSR is applied to both image and 3D model texture map SR where multiple HR reference images are available. The proposed method is evaluated on benchmark datasets and demonstrated to significantly improve performances. 
We introduce 3 new multiple reference SR datasets which will be made available to benchmark future SR approaches. 
\section{Related work}
\subsection{Single-image super-resolution (SISR)}
\noindent
A breakthrough in the SISR task was achieved when Dong \etal~\cite{dong2014learning} tackled the problem with a convolutional neural network (CNN). From this work, the application of deep learning progressively replaced classic SR computer vision methods~\cite{wang2020deep}. The pioneer work of Dong \etal~\cite{dong2014learning} belongs to a group of SR methods that use mean squared error (MSE) as their objective function. VDSR~\cite{kim2016accurate} shows the importance of a deep layer architecture while SRResNet~\cite{ledig2017photo} and EDSR~\cite{lim2017enhanced} demonstrate the benefit of using residual block~\cite{he2016deep} to alleviate the training. Several modifications of the residual structure such as skip connections~\cite{tong2017image}, recursive structures~\cite{tai2017image} and channel-attention~\cite{zhang2018image} further improved the accuracy of SISR. The state-of-the-art CSNLN~\cite{mei2020image} integrates a cross-scale non-local attention module to learn dependencies between the LR and HR images. Other works propose lightweight networks to alleviate computational cost~\cite{muqeet2020ultra,liu2020residual}.
These residual networks ignore the human perception and only aim to high values of PSNR and SSIM, producing blurry SR images~\cite{wang2020deep}. Generative adversarial networks (GANs), introduced in the SR task by Ledig \etal with SRGAN~\cite{ledig2017photo}, aim to enhance the perceptual quality of the SR images.
The performances of SRGAN were improved by ESRGAN~\cite{ledig2017photo}, which replaces the adversarial loss with a relativistic adversarial loss. RSRGAN~\cite{zhang2019ranksrgan} develops a rank-content loss by training a ranker to obtain state-of-the-art visual results. 
%
\\\indent \textbf{3D appearance super-resolution: }There are only two deep learning works that super-resolve texture maps to enhance the appearance of 3D objects. 
The method proposed by Li \etal~\cite{li2019_3dappearance} processes, with a modified version of EDSR~\cite{lim2017enhanced}, LR texture maps and their normal maps to incorporate geometric information of the model in the learning. The pre-process to create normal maps introduces heavy computational cost. 
In the second work~\cite{richard2019learned}, a redundancy-based encoder generates a blurry texture map from LR images that is then deblurred by a SISR decoder. Its main objective is not the super-resolution but the creation of texture maps from a set of LR multi-view images.
\subsection{Reference-based super-resolution (RefSR)}
\noindent
GANs were introduced to solve the problems of the residual networks by focusing on the perceptual quality of the image. However, their generative nature leads to the creation of unnatural textures in the SR image. RefSR approaches were applied to eliminate these artefacts by learning more accurate details from reference images.
One of the first RefSR networks is CrossNet~\cite{zheng2018crossnet}, which uses optical flow to align input and reference, limiting the matching of long distance correspondences. CrossNet was improved with two-stage cross-scale warping modules, adding to the optical flow alignment a further warping stage~\cite{tan2020crossnet++}. The optical flow introduces artefacts when misaligned references must be handled.
The ``patch-match" approach correlates the reference and input images by matching similar patches. In an early, non deep learning framework~\cite{boominathan2014improving}, patches of downsampled reference image are matched with gradient features of the LR image. This work was adapted by Zheng \etal~\cite{zheng2017learning} to perform semantic matching as well as to synthesise SR features through a CNN. More recently, Zhang \etal proposed SRNTT~\cite{ZhangSRNTT2019}, which swaps the most similar features of the reference and the LR image through convolutional layers. TTSR~\cite{yang2020learning} refines the matching mechanism by selecting and transferring only relevant textures from the reference image. SSEN~\cite{shim2020robust} performs the patch-match through deformable convolution layers using an offset estimator. MASA~\cite{lu2021masa} adds a spatial adaptation module to handle large disparity in color or luminance distribution between reference and input images.
CIMR~\cite{cui2020towards} is the only method in the literature that exploits multiple references. It selects a smaller subset from all the features of generic reference images without performing any comparison with the LR input, neglecting similar textures of the references.
%
Our approach utilises a hierarchical patch-match method to search for relevant textures among all the feature vectors of multiple references, increasing the possibility to find more similar high-quality textures. It performs an attention-based similarity mapping between the references and subvectors of the LR input, improving performances. Finally, it significantly reduces the GPU usage of the patch-match approaches, facilitating the reproducibility of RefSR studies.
%
\section{Adaptive multi-reference super-resolution}
%
%
\noindent
In this section we present the proposed AMRSR network, designed to exploit multiple HR reference images for training and inference whilst maintaining memory efficiency. A hierarchical attention-based approach is introduced for image feature matching from the LR input to the HR reference images using a perceptual loss.   
%
Hierarchical attention allows multiple HR reference images without a significant increase in GPU memory requirements and is demonstrated to improve performance versus a global similarity search (section \ref{ssec:gpu}).  
The problem of multiple-reference super-resolution can be stated as follows:
given a LR input $I_{LR}$ and a set of HR reference images $\{I^m_{ref}\}_{m=1}^{N_{M}}$, estimate a spatially coherent SR output $I_{SR}$ with the structure of $I_{LR}$ and the appearance detail resolution of the multiple-reference images.   
The SR output should contain perceptually plausible HR appearance detail without the introduction of visual artefacts such as blur, ringing or unnatural discontinuities observed with previous SR approaches for large up-scaling factors ($>2\times$).
\subsection{Overview of approach}
\begin{figure}[!]
    \centering
    \resizebox{1\linewidth}{!}{
    \includegraphics[width=0.85in]{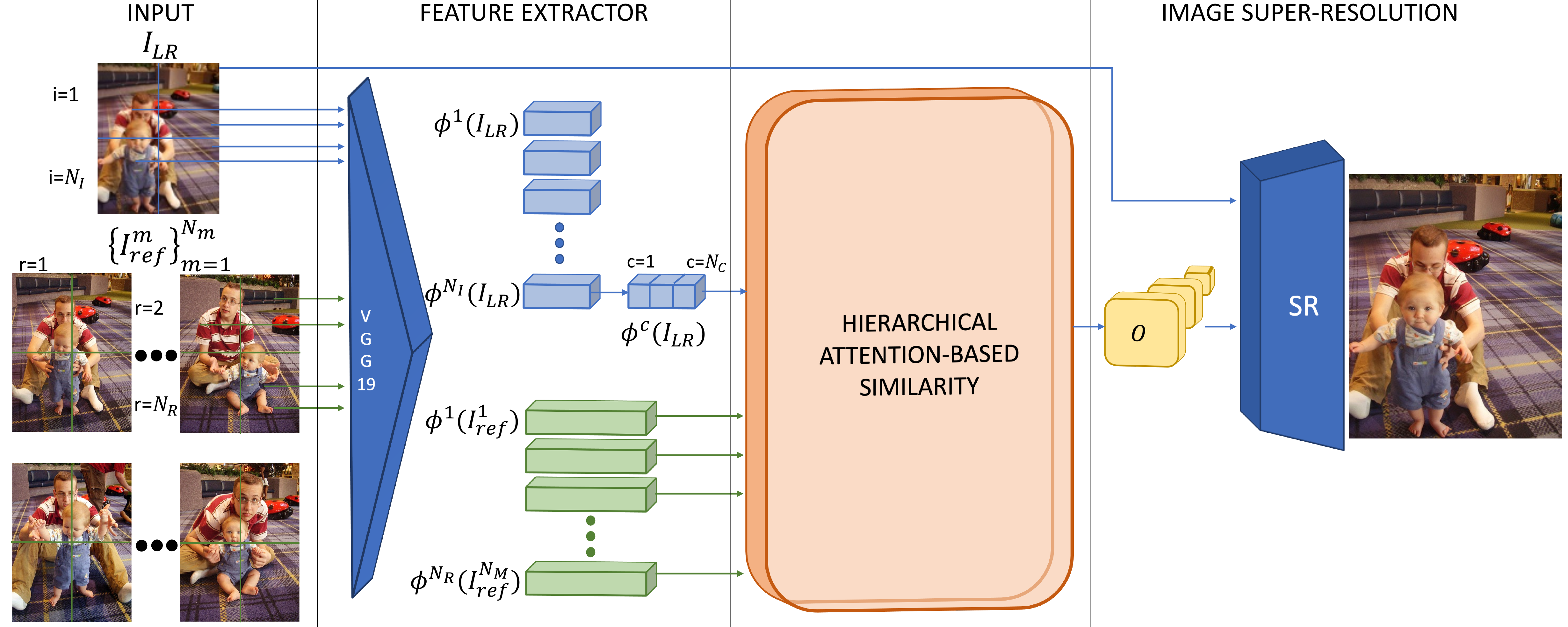}}
    \vspace{-2.0em}
    \caption{AMRSR inference overview. Given a LR input image and $N_M$ reference images, AMRSR comprises three modules: feature extraction; hierarchical attention-based similarity; and image SR sampling. The resulting output is a HR reconstruction of the LR input image.}
    \vspace{-1.8em}
    \label{fig:network}
\end{figure}
\noindent
Figure \ref{fig:network} presents an overview of the proposed approach, which comprises the following stages.
\\
\textbf{Feature Extraction:} to reduce GPU memory consumption with multiple reference images, the LR input $I_{LR}$ and HR reference images $\{I^m_{ref}\}^{N_M}_{m=1}$ are divided into $N_I$ and $N_{R}$ sub-parts, respectively. Image features are extracted from these parts using a pre-trained VGG-19 network~\cite{simonyan2014very}. 
\\
\textbf{Hierarchical Attention-based Similarity:} computes a mapping of features from the LR image to the most similar features of the HR reference images. The similarity $s_k$ is inferred between the feature vector of the LR input and of every reference image. The multiple references are then sampled based on the most similar features. This process is executed following a hierarchical structure of $l=N_L$ levels with an attention-based similarity mapping of the input feature vector. The output $O$ is a feature vector containing the most similar reference features to the input features.
\\
\textbf{Image Super-resolution:} given the feature similarity mapping $O$, a convolutional network super-resolves the LR input $I_{LR}$ to obtain the SR output $I_{SR}$ which maintains the spatial coherence of the input with the HR appearance detail of the reference images.
\begin{figure}[!]
    \centering
    \resizebox{0.85\linewidth}{!}{
    \includegraphics[width=1.in]{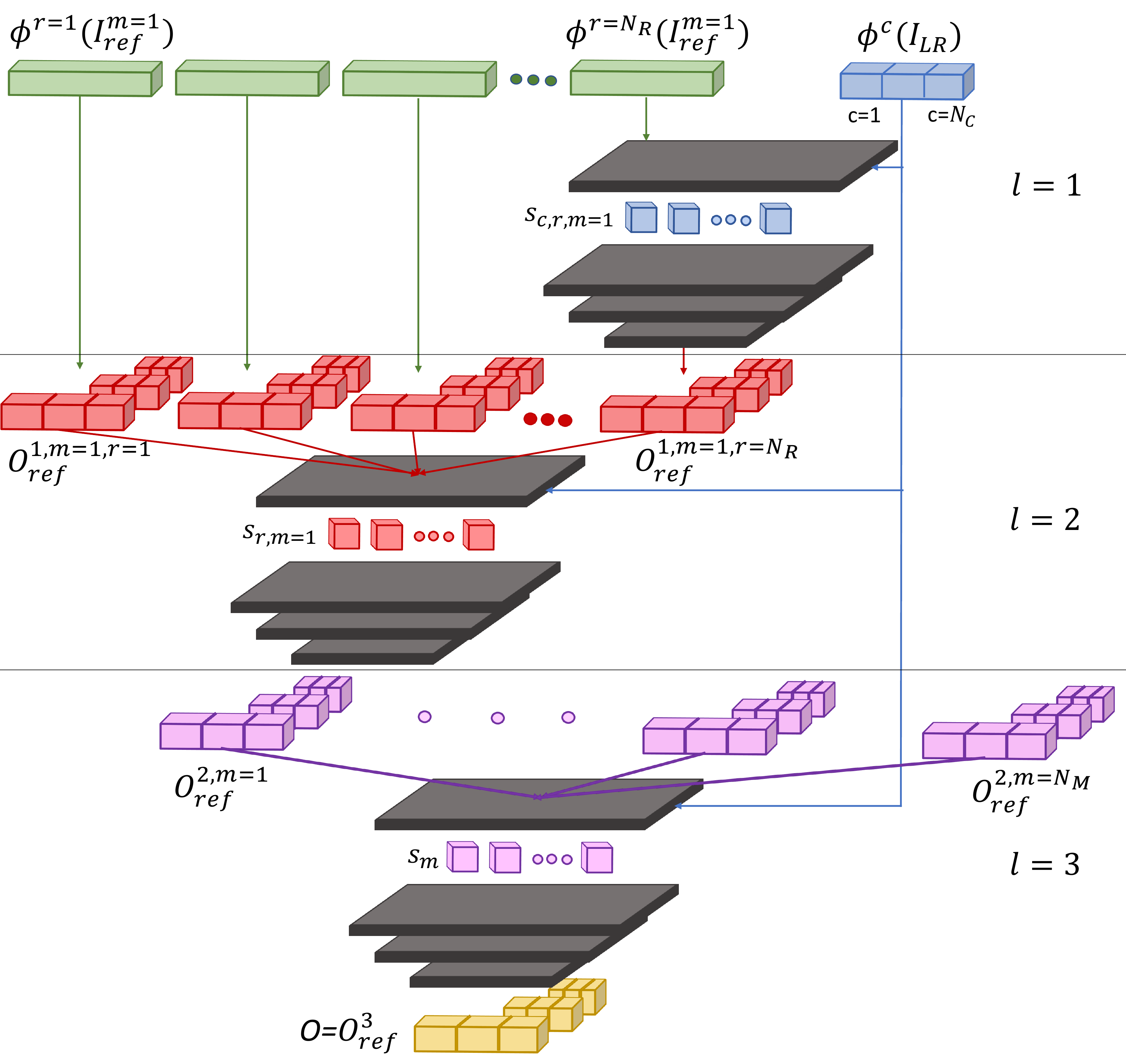}}
    \vspace{-1.em}
    \caption{Hierarchical attention-based similarity with $N_L=3$ levels.} 
    \vspace{-1.8em}
    \label{fig:network2}
\end{figure}
\noindent
\\In contrast to the patch-match adopted by previous RefSR approaches \cite{ZhangSRNTT2019,yang2020learning,li2020survey,shim2020robust,lu2021masa}, AMRSR performs feature similarity matching between subvectors of the LR input and reference images to focus the learning attention on the input features. A feature vector contains $N$ features of the input image, each of which is a matrix of values that represents a specific image feature. Instead of processing the whole matrix, we divide it into submatrices and perform the similarity mapping with these. 
This improves the learning of the most similar features in each submatrix giving improved performance (Table~\ref{tbl:attention}). We refer the reader to the supplementary material (Table 2) for an outline of the notation. 
\subsection{Feature extraction}
\noindent
To conduct the similarity matching in the neural domain, feature vectors of the LR input and references must be retrieved. A problem of previous RefSR approaches is that the patch-match on HR reference images requires high GPU memory usage. To tackle this, we divide the input and reference images into $N_I$ and $N_{R}$ sub-parts, respectively. Feature vectors of these parts are extracted with a VGG-19 network~\cite{simonyan2014very} as shown in the second part of Figure~\ref{fig:network}. The feature vectors of each part of the input are divided into $N_C$ subvectors $\{\{\phi^c_i(I_{LR})\}_{i=1}^{N_I}\}_{c=1}^{N_C}$ to perform an attention-based similarity on the LR input. For simplicity, we assume $N_I=1$ (the input image is not divided into parts) and we express the set as $\phi^c(I_{LR})$ without loss of generality. If $N_I>1$, the algorithm is repeated for each part and the outputs are concatenated. $N_M \times N_R$ feature vectors are retrieved for the references $\{\{\phi^r(I_{ref}^m)\}_{m=1}^{N_M}\}_{r=1}^{N_{R}}$ (expressed as $\phi^r(I_{ref}^m)$). Dividing the input and reference images into parts and inferring the similarity between them in a hierarchical order, establishes an efficient mechanism that improves performances and reduces GPU memory requirements. This is important for practical implementation of multi-reference SR within a fixed GPU memory size.
\subsection{Hierarchical attention-based similarity}
\noindent
The objective of this stage is to map the features of the LR input to the most similar features of the HR reference images. The output is a feature vector that contains the values of these most similar reference features. 
A hierarchical approach of similarity mapping is performed over $l=N_L$ levels.
For every level $l$ of the hierarchy, a similarity map between LR input subvectors and reference features is computed. The most similar features are then retrieved considering the maximum values of the similarity map. A new feature vector is created with these features and used to compute the similarity map and the feature vector in the next level of the hierarchy.
\\
The similarity map $s^l_k$ for level $l$ is evaluated by convolution between the subvectors $\phi^c(I_{LR})$ of the LR input and $O^{l-1,r,m}_{ref}$, which is either the input reference feature vectors $\phi^r(I^m_{ref})$ if $l=1$ or new vectors created in the level $l=l-1$ (which contain features of the references $\{I^m_{ref}\}$):
\begin{equation}
\label{eq:1}
    s^{l}_{k}=\phi^c(I_{LR})*\frac {P_{k}(O^{l-1,r,m}_{ref})}{||P_{k}(O^{l-1,r,m}_{ref})||} \tag{1}
\end{equation}
$k=c$ if $l=1$, $k=r$ or $k=m$ otherwise. $P$ is the patch derived from the application of the patch-match approach: patches of $O^{l-1,r,m}_{ref}$ are convoluted with $\phi^c(I_{LR})$ to compute the similarity. 
\\
When the similarity map $s^{l}_{k}$ is evaluated, a vector $O^{l}_{ref}$ containing the most similar features of $O^{l-1}_{ref}$ is created by applying either one of two distinct approaches:
\begin{enumerate}[noitemsep]
    \item \textbf{Input attention mapping} ($l=1$): in the first level a feature vector is created by maximising over every subvector of the input:
    \begin{gather}
        \label{eq:2}
        O^{1,r,m}_{ref}(x,y)=P_{k^*}(\phi^r(I^m_{ref}))(x,y) \tag{2}\\ k^*=\argmax_{k=c}s^{1}_{k}(x,y)\notag 
    \end{gather}
    $O^{1,r,m}_{ref}(x,y)$ represents a single value in the $(x,y)$ position of the created feature vector $O^{1,r,m}_{ref}$. This value corresponds to the $(x,y)$ value of the $k^*$ patch $P(\phi^r(I^m_{ref}))$ whose $s^1$ is the highest among all the similarity values $s^{1}_{k}(x,y)$ for each subvector of the LR input feature vector.
    \item \textbf{Reference attention mapping ($l>1$)}: for subsequent levels of the hierarchy, a feature vector is created by maximising a new similarity $s^{l}_{k}$ map over the feature vector created in the previous level.
    \begin{gather}
        \label{eq:3}
        O^{l,k}_{ref}(x,y)=O^{l-1,k^*}_{ref}(x,y) \tag{3}\\ 
        k^*=\argmax_{k}s^{l}_{k}(x,y)\notag 
    \end{gather}
    $k=r$ or $ k=m$ depending on which level is processed. The value of $O^{l,k}_{ref}$ in the $(x,y)$ position is the value of $O^{l-1,k}_{ref}$ with the highest $s^{l}$ among all the $s^{l}_{k}(x,y)$ of  $O^{l-1,k}_{ref}$.
\end{enumerate}
Mapping is repeated at multi-scales with three feature extractor levels to achieve robustness to the variance of colour and illumination~\cite{ZhangSRNTT2019}.
The final output, obtained when the similarity mapping is performed for all the levels of the hierarchy, is a feature vector $O=O^{N_L}_{ref}$ which contains the features of the reference images that are most similar to every feature of the LR input. 
When the final level of the hierarchy is processed, $N_K$ sets of weights $W_k$ are computed as the maximum of the scalar product between $\phi^c(I_{LR})$ and the $O^{l-1,k}_{ref}$ vector produced in the previous level.
\begin{equation}
\label{eq:4}
    W_k=max(\phi^c(I_{LR}) \cdot O^{l-1,k}_{ref})\tag{4}
\end{equation}
The final set of weights $W$ is then retrieved from these sets: the weight in position $(x,y)$ of $W$ has the same value of the weight in position $(x,y)$ of the $k^*$-th set $W_{k^*}$ with $k^*$ from Equation~\ref{eq:3} with $l=N_L$: $W(x,y)=W_{k^*}(x,y)$.
\\
For $N_L=3$ levels of hierarchy as shown in Figure~\ref{fig:network2}:
   \\ $\bf l=1$: feature similarity mapping between every subvector of the input vector and every part of every reference. \textit{Input:} $\phi^c(I_{LR})$, $\phi^r(I_{ref}^m)$, $k=c$. \textit{Output:} $\{\{O^{1,m,r}_{ref}\}_{m=1}^{N_{M}}\}_{r=1}^{N_{R}}$.
    \\$\bf l=2$: feature similarity mapping between the input subvectors and all the $N_R$ parts of a single reference, repeated for every reference. \textit{Input:} $\phi^c(I_{LR})$, $\{\{O^{1,m,r}_{ref}\}_{m=1}^{N_{M}}\}_{r=1}^{N_{R}}$, $k=r$.  \textit{Output:} $\{O^{2,m}_{ref}\}_{m=1}^{N_{M}}$.
    \\$\bf l=3$: feature similarity mapping between the input subvectors and all the references. \textit{Input:} $\phi^c(I_{LR})$, $\{O^{2,m}_{ref}\}_{m=1}^{N_{M}}$, $k=m$. \textit{Output}: $O=O^3_{ref}$.
\subsection{Image super-resolution} 
\noindent
In the last stage, $I_{LR}$ is super-resolved with a generative network that exploits the information of the vectors obtained with the hierarchical similarity mapping whilst maintaining the spatial coherence of $I_{LR}$. These vectors are embedded to the input feature vector through channel-wise concatenation in different layers $h$ of the network.
We modified the architecture of the generator used by Zhang \etal~\cite{ZhangSRNTT2019} by eliminating the batch normalization layers since they can reduce the accuracy for dense pixel value predictions~\cite{yu2018wide}. More details are explained in the supplementary material.
A texture loss is defined to enforce the effect of the texture swapping between $I_{LR}$ and the obtained $O$:
\begin{equation}
\label{eq:5}
    L_{tex}=\sum_{h}||Gr(\phi_h(I_{SR}\cdot W_h))-Gr(O_h\cdot W_h))|| \tag{5}
\end{equation}
where $Gr(\cdot)$ computes the Gram matrix.
Differently from~\cite{ZhangSRNTT2019}, the weighting map $W_h$ is computed among the $N_M$ references. The weight of HR image features more similar to $I_{LR}$ will be higher. Thus the appearance transfer from $\{I_{ref}\}$ to $I_{SR}$ is adaptively enforced based on the references similarity. In addition, the network minimises:
\begin{itemize}[noitemsep]
\item The adversarial loss $L_{adv}$ to enhance the visual quality of the SR output. To stabilize the training, we use the WGAN-GP~\cite{gulrajani2017improved} for its gradient penalization feature.
\item The reconstruction $L_1$ loss, since it has been demonstrated to give sharper performance than $L_2$ loss~\cite{yang2020learning}.
\item The perceptual loss~\cite{johnson2016perceptual} $L_{p}$ to enhance the similarity between the prediction and the target in feature space. 
\end{itemize}
\section{Dataset}
\label{sec:dataset}
\noindent
To the best of our knowledge, no multi-reference benchmark datasets are available (only with a single reference~\cite{ZhangSRNTT2019}). 
To achieve our objective of multi-reference SR, we introduce three datasets:
\vspace{0.05in}
\\\textbf{1. CU4REF}: this dataset is built from the single reference dataset CUFED5~\cite{ZhangSRNTT2019}. 4 groups of images are defined from the CUFED dataset~\cite{wang2016event}, each with a different similarity level from the LR input images. We use the images in these groups as our references. The training set contains 3957 groups of LR and reference images while the testing set contains 126 groups (4 references for every LR image).
\\\textbf{2. HUMAP:} to create the references of 67 synthetic human texture maps downloaded from several websites \cite{renderpeople,cgtrader,turbosquid,free3d}, we import their 3D models in Blender~\cite{blender} and render 8 camera views as reference images for each subject. Two real human texture maps retrieved from 16 multi-view images are added. Due to the low amount of data, we augment the dataset by cropping the texture maps into patches (256x256 size)~\cite{li2019_3dappearance}. 
The training dataset consists of 5505 groups of patches and references. The testing dataset comprises 336 groups created from 5 texture maps of 6 subjects (3 captured by 16 video-cameras, 2 using a 5x5 FLIR Grasshooper3 camera array) and a texture map of 2 people~\cite{collet2015high}. 
\\\textbf{3. GEMAP:} consists of generic LR texture maps associated with 8 references. The texture maps are taken from the 3DASR dataset~\cite{li2019_3dappearance}, created from the multi-view images and 3D point clouds from other datasets (\cite{ley2016syb3r},\cite{seitz2006comparison},\cite{schops2017multi},\cite{goldlucke2014super},\cite{zhou2014color},\cite{zollhofer2015shading}). The reference images for the texture maps of~\cite{ley2016syb3r} are created with the same approach applied for HUMAP. For the other texture maps, the HR multi-view images captured by DSLR cameras are taken as references. The LR texture maps are cropped as in HUMAP. The training dataset contains 2032 groups and for testing 290 groups.
    \vspace{-0.7em}
\\To evaluate the generalization capability of AMRSR on RGB images, we test it on Sun80 dataset~\cite{sun2012super}, which has 80 natural images accompanied by a series of web-search references that significantly differ from the input images.
\section{Results and evaluation}
\label{sec:exp}
\noindent
This section illustrates how AMRSR outperforms other state-of-the-art methods with quantitative and qualitative comparisons. Two ablation studies on the network configuration and on the advantage of multiple references are then presented. The GPU memory requirement of the state-of-the-art RefSR approaches is compared confirming the efficiency of AMRSR. The LR inputs are obtained by bicubic downscaling ($4\times$) from their ground-truth HR images and the SR results are evaluated on PSNR and SSIM on the Y channel of YCbCr space. AMRSR parameters are: $N_M=4$, $N_I=1$, $N_R=1$ for CU4REF dataset, $N_R=16$ for all others (see Section~\ref{ssec:gpu}). To integrate the input subvectors with the architecture structure, the value of $N_C$ is:
\begin{equation}
    \label{eq:6}
    N_C(x)=\frac{length(\phi(I_{LR})(relu3\_1))}{length(\phi(I_{LR})(reluq\_1)_{q=1}^{3})/4}\tag{6}
\end{equation}
where $q$ indicates the three different layers used for the multi-scale fashion approach.
Further evaluations and results are presented in the supplementary material.
\subsection{Comparison with state-of-the-art approaches}
\begin{table}[!]
 \centering
\resizebox{0.45\textwidth}{!}{\input{amrsr_arxive/table/SR2_new_v2}}
\vspace{-0.8em}
\caption{PSNR/SSIM values of different SR approaches. * indicates that the references are downscaled of a factor of 2 (see Section\ref{ssec:gpu}).}
\vspace{-1.7em}
\label{tbl:quant}
\end{table}
\begin{table*}[!h]
\centering
\resizebox{0.85\linewidth}{!}{\input{amrsr_arxive/image_files/comparison}}
 \vspace{-1.1em}
 \captionof{figure}{Qualitative comparison among the state-of-the-art SR approaches on CU4REF (first two examples), Sun80 (third and fourth), GEMAP (fifth) and HUMAP (last). The references are shown for each example. The top left or the most left reference was used in the single-reference SR approaches.}
 \vspace{-1.8em}
 \label{fig:qualit}
\end{table*}
\noindent 
Qualitative and quantitative comparisons are performed with state-of-the-art SISR and RefSR approaches. The SISR methods are the PSNR-oriented EDSR~\cite{lim2017enhanced}, MDSR~\cite{lim2017enhanced}, RRDBNet~\cite{wang2018esrgan}, SRResNet~\cite{ledig2017photo}, RCAN~\cite{zhang2018image}, NHR~\cite{li2019_3dappearance}, NLR~\cite{li2019_3dappearance}, CSNLN~\cite{mei2020image}, MAFFSRN~\cite{muqeet2020ultra} and the visual-oriented SRGAN~\cite{ledig2017photo}, ESRGAN~\cite{wang2018esrgan}, RSRGAN~\cite{zhang2019ranksrgan}. The RefSR approaches are CrossNet~\cite{zheng2018crossnet}, SSEN~\cite{shim2020robust}, SRNTT~\cite{ZhangSRNTT2019}, TTSR~\cite{yang2020learning} and MASA~\cite{lu2021masa} (published June '21).
We train each network with the datasets presented in Section~\ref{sec:dataset} with the same training configurations. Training with adversarial loss usually deteriorates the quantitative results. For a fair comparison with the PSNR-oriented methods, we train our model, SRNTT, MASA and TTSR only on reconstruction loss (named with the suffix ``\textit{$l_2$}").
NHR and NLR are tested on HUMAP and GEMAP since they require normal maps (retrieved with Blender~\cite{blender}).
\\\indent\textbf{Quantitative comparison: }the PSNR and SSIM values of each method are presented in Table~\ref{tbl:quant}, which is divided into two parts: PSNR-oriented networks in the upper part; visual-oriented GANs and RefSR in the lower part. The highest scores are highlighted in red while the second highest scores are blue. The bold red figures are the highest across both PSNR- and visual-oriented methods. AMRSR-\textit{$l_2$} achieves the highest values of PSNR and SSIM in the PSNR- and visual-oriented methods for all four datasets.
\\\indent\textbf{Qualitative comparison: }
Figure~\ref{fig:qualit} shows SR examples of the most relevant approaches considered in our evaluation. The SR outputs produced by the PSNR-oriented methods (RCAN, CSNLN) are blurrier and the details are less sharp. The results produced by the visual-oriented approaches (RSRGAN, SRNTT, TTSR) present unpleasant artefacts such as ringing and unnatural discontinuities.
The SR outputs of AMRSR are less blurry with sharper and finer details as shown in the zoomed patches of the examples, whose quality is higher than the other results.
\\\indent\textbf{User study evaluation: }
to further evaluate the visual quality of the SR outputs, we conduct a user study, where AMRSR is compared with five approaches.
50 random SR output pairs were shown to 60 subjects. Each pair consists of an image of AMRSR and the counterpart generated by one of the other approaches. The users were asked to pick the image with the best visual quality. The values on the Y-axis of Figure~\ref{fig:chart} illustrate the percentage of the users that select AMRSR outputs. AMRSR significantly outperforms the other methods with over 90\% of users voting for it.
\begin{figure}[!]
\centering
\resizebox{0.8\linewidth}{!}{\input{amrsr_arxive/image_files/chart}}
\vspace{-1.em}
\caption{Percentage of users that prefer AMRSR over 5 approaches in the user study. The error bars indicate the $95\%$ confidence interval.}
\vspace{-1.em}
\label{fig:chart}
\end{figure}
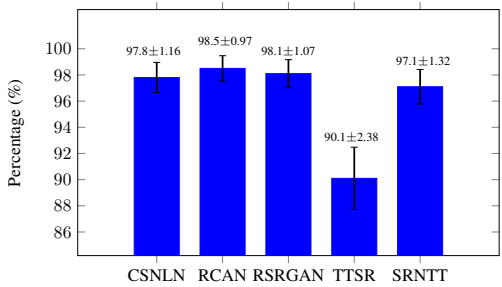
\begin{table}[!]
    \centering
    \resizebox{0.45\textwidth}{!}{\input{amrsr_arxive/table/simlevel_v2}}
    \vspace{-0.8em}
    \caption{Quantitative comparison between AMRSR and RefSR methods using single reference with different levels of similarity to the LR input.}
     \vspace{-2.0em}
    \label{tab:difref}
\end{table}
\\\indent\textbf{Influence of dissimilar references: }similarity between LR and reference images significantly influences the performances of RefSR methods~\cite{ZhangSRNTT2019}. Following the setting in~\cite{ZhangSRNTT2019}, we evaluate the effect of dissimilar reference images to the LR input testing on four similarity levels (from L1 to L4) of the CUFED5 test set, defined by computing the number of best matches of SIFT features between the input image and the references. The references that belong to L1 have the highest number of matches while the L4 ones have the lowest number. The highest figures of PSNR and SSIM (Table~\ref{tab:difref}) are obtained by our network when a single reference is used even though its level of similarity has decreased, confirming the higher efficiency of AMRSR when the references are not similar to the LR input. To show the efficiency of leveraging multiple references, we evaluate the RefSR approaches on the CU4REF dataset by swapping the references of an image with other images in the dataset. The quantitative results of Table~\ref{fig:aim} and the visual comparison in Figure~\ref{tab:sim} demonstrate the benefits of using multiple dissimilar references. AMRSR is able to find more similar patches within multiple references even if these are very dissimilar to the LR input. For dissimilar references, AMRSR achieves higher PSNR and SSIM than other methods and than when a single reference is used (AMRSR 1). This study demonstrates the performance of adaptive sampling of AMRSR even for references with dissimilar features.
\\\indent\textbf{Comparison with CIMR~\cite{cui2020towards}: } we compare AMRSR with CIMR~\cite{cui2020towards}, the only other multi-reference super-resolution approach. AMRSR is trained following the setting in~\cite{cui2020towards} on CUFED5 dataset (13,761 images). We randomly associated $N_M$ references taken from Outdoor Scene (OST) dataset~\cite{wang2018recovering} to each LR input image.
CIMR is evaluated on content-independent references using a reference pool to select a subset of feature vectors from 300 reference images. We evaluate AMRSR by randomly associating to the LR input images $N_M$ reference images from the 300 images.
Results presented in Table~\ref{tbl:cimr2} show that AMRSR outperforms CIMR in the multiple references case.
\begin{table}[!]
    \centering
    \resizebox{0.48\textwidth}{!}{\input{amrsr_arxive/table/sim2_v2}}
    \vspace{-1.0em}
    \caption{PSNR/SSIM values with references dissimilar to the LR input.}
    \vspace{-1.0em}
    \label{tab:sim}
\end{table}
\begin{table}[!]
\centering
\resizebox{0.9\linewidth}{!}{\input{amrsr_arxive/image_files/sim}}
 \vspace{-1.2em}
 \captionof{figure}{Visual comparison of SR outputs obtained with multiple references dissimilar to the LR input.}
 \vspace{-1.0em}
 \label{fig:aim}
\end{table}
\begin{table}[!]
    \centering
    \resizebox{0.45\textwidth}{!}{\input{amrsr_arxive/table/cimr2}}
    \vspace{-0.8em}
    \caption{Quantitative comparison between AMRSR and CIMR exploiting multiple reference images. The results of CIMR are taken from~\cite{cui2020towards}.}
    \vspace{-1.7em}
    \label{tbl:cimr2}
\end{table}
\subsection{Ablation studies}
\label{ssec:gpu}
\indent\textbf{Number of reference images: }a key contribution of our work is the transfer of high-quality textures from multiple references to increase the matching between similar LR input and HR reference features.
To prove this, we test AMRSR by changing the number of references. The results of using $N_M=1, 2, 4$ reference images are compared for CU4REF and Sun80 datasets. $N_M=8$ references are also considered for HUMAP and GEMAP. Table~\ref{tbl:ref} presents the PSNR and SSIM figures for the different cases, including the second best results of related works (``$2^{nd}$ best").
Increasing the number of references leads to higher values of PSNR and SSIM. The highest ones are generally obtained with the maximum number of references. AMRSR outperforms the $2^{nd}$ best techniques also when a single reference is used. 
Figure~\ref{fig:refere} confirms the advantage of using multiple references. The hairs of the human subject in the texture map example are sharper when 4 references are used because they are transferred from the side of the model, which is not visible with 1 or 2 references. 
Similarly, the window grates are neat only when 4 references are used because the network learns their texture from the third reference.
\indent\textbf{Attention mapping: }
we evaluate the effect of processing subvectors of the LR input in the similarity mapping by comparing the actual configuration of AMRSR with three others: (i) without any attention mapping (processing the whole input vector); (ii) with an attention mapping in the reference feature vectors (dividing them into subvectors); (iii) with an attention mapping in both the input and reference feature vectors.
The obtained values of PSNR and SSIM are shown in Table~\ref{tbl:attention}.
The effectiveness of the input attention mapping is proved by both the quantitative and qualitative evaluation, with a significant boost in the performance and higher quality of the SR examples of Figure~\ref{fig:atten}.
\begin{table}[!]
 \centering
\resizebox{0.45\textwidth}{!}{\input{amrsr_arxive/table/reference_psnr2}}
\vspace{-0.8em}
\caption{PSNR/SSIM values obtained with different numbers of reference images. The bottom row shows the figures for the second-best approaches.}
\vspace{-1.0em}
\label{tbl:ref}
\end{table}
\begin{table}[!]
 \centering
\resizebox{0.45\textwidth}{!}{\input{amrsr_arxive/table/attention_psnr2}}
\vspace{-0.8em}
\caption{PSNR/SSIM values of different configurations of AMRSR obtained by dividing into subvectors the feature vectors of references (ref), of both reference and input (both) or none (no).}
\vspace{-1.0em}
\label{tbl:attention}
\end{table}
\begin{table}[!]
\centering
\resizebox{1.\linewidth}{!}{

\input{amrsr_arxive/image_files/attention}}
 \vspace{-1.2em}
 \captionof{figure}{Visual comparison of the results obtained by changing attention.}
 \vspace{-2.0em}
 \label{fig:atten}
\end{table}
\\\indent\textbf{Part-based mechanism and GPU memory usage: }
we finally evaluate the effect of the part-based mechanism of AMRSR by computing the required GPU memory of a Quadro RTX 8000 GPU (48 GB) during inference and comparing it with TTSR, SRNTT, MASA, AMRSR with different numbers of references and with different values of $N_{R}$ (Table~\ref{tbl:memory}). Changing the value of $N_{R}$ leads to a modification of the hierarchical structure: if $N_{R}=1$, the $2^{nd}$ level of the hierarchy is skipped. The maximum size of the reference images is: (500x500) for CU4REF, (1024x928) for Sun80, (5184x3840) for HUMAP and (4032x6048) for GEMAP.
Increase in the reference image size results in increased memory consumption and improve super-resolution quality. AMRSR requires significantly less memory than the single RefSR approaches when multiple references are used. To test TTSR, SRNTT and MASA with HUMAP and GEMAP datasets, the reference must be downscaled ($2\times$) to consume less than 48GB. For a fair comparison, we test our network with 1 and 4 downscaled references (``cut" in the table). The comparison between different choices of $N_{R}$ shows that, when higher resolution references are exploited with higher values of $N_{R}$, the memory footprint is reduced and the PSNR and SSIM are increased. For CU4REF, the best performances are achieved with $N_{R}=1$ because the size of its references is much lower. 
\section{Conclusion}
\begin{table}[!]
\centering
\resizebox{1\linewidth}{!}{\input{amrsr_arxive/image_files/reference2}}
 \vspace{-1.2em}
 \captionof{figure}{Visual comparison of the results obtained by changing the number of references. The first example is of HUMAP, the other is of Sun80.}
 \vspace{-1.0em}
 \label{fig:refere}
\end{table}
\begin{table}[!]
 \centering
\resizebox{0.48\textwidth}{!}{\input{amrsr_arxive/table/memory3_v2}}
\vspace{-0.8em}
\caption{PSNR/SSIM values and GPU memory usage (in GB) of different configurations of AMRSR and the state-of-the-art RefSR approaches. $N_{R}$ is the number of parts which the reference images are divided into.}
\vspace{-2em}
\label{tbl:memory}
\end{table}
\noindent
In this paper, we tackle the super-resolution problem with a multiple-reference super-resolution network that is able to transfer more plausible textures from several references to the super-resolution output. Our network focuses the learning attention in the comparison between subvectors of the low-resolution input and the reference feature vectors, achieving significant qualitative and quantitative improvements as demonstrated from the evaluation. A hierarchical part-based mechanism is introduced to reduce the GPU memory usage, which is prohibitive if previous RefSR methods are applied with high-resolution reference images. In addition, we introduce 3 datasets to facilitate the research for multiple-reference and 3D appearance super-resolution.
\\
\noindent{\bf Acknowledgement:} This research was supported by UKRI EPSRC Platform Grant EP/P022529/1. 

{\small
\bibliographystyle{ieee_fullname}
\bibliography{egbib}
}

\end{document}

%% file: amrsr_arxive/image_files/intro2.tex
\begin{huge}
\begin{tabular}{cccc}
            Ref. images & Ground Truth & GT patch & LR patch \\
             \includegraphics[width=2in]{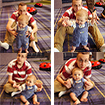} & \includegraphics[width=2in]{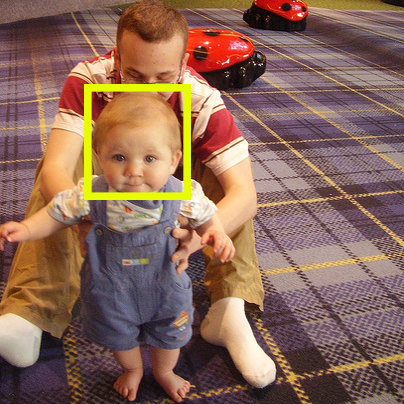} & \includegraphics[width=2in]{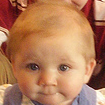} & \includegraphics[width=2in]{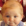} \\
               CSNLN~\cite{mei2020image}& RSRGAN~\cite{zhang2019ranksrgan}&SRNTT~\cite{ZhangSRNTT2019}&AMRSR(ours)\\
             \includegraphics[width=2in]{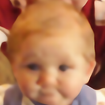} & \includegraphics[width=2in]{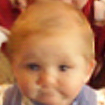}& \includegraphics[width=2in]{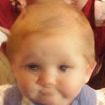} &
             \includegraphics[width=2in]{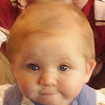}  \\
        \end{tabular}
        \end{huge}

%% file: amrsr_arxive/table/SR2_new_v2.tex
\begin{tabular}{c|c|cccc|}
\cline{2-6}
&
\multicolumn{1}{c}{\textbf{Methods}}   & \multicolumn{1}{c|}{\textbf{CU4REF}} & \multicolumn{1}{c|}{\textbf{Sun80}}                    & \multicolumn{1}{c|}{\textbf{GEMAP}} & \multicolumn{1}{c|}{\textbf{HUMAP}}                      \\ \hline
\multicolumn{1}{|c|}{\multirow{12}{*}{\rotatebox[origin=c]{90}{\textbf{\textit{PSNR-oriented}}}}} & SRResNet~\cite{ledig2017photo}  & 26.28/.7823                          & 29.80/.8121                                            & 35.71/.9093                         & 46.06/.9785                                              \\
\multicolumn{1}{|c|}{}                        &
RRDBNet~\cite{wang2018esrgan}   & 26.22/.7828                          & 29.56/.8053                                            & 35.77/.9102                         & 46.25/.9790                                              \\
\multicolumn{1}{|c|}{}                        &
EDSR~\cite{lim2017enhanced}      & 25.52/.7652                          & 28.74/.7876                                            & 35.36/.9051                         & 45.92/.9784                                              \\
\multicolumn{1}{|c|}{}                        &
MDSR~\cite{lim2017enhanced}      & 26.43/.7822                          & 29.96/.8137                                            & 35.84/.9107                         & 46.06/.9784                                              \\
\multicolumn{1}{|c|}{}                        &
NHR~\cite{li2019_3dappearance}       &            --                          &    --                                                    & 33.13/.8981                         & 36.15/.9544                                              \\
\multicolumn{1}{|c|}{}                        &
NLR~\cite{li2019_3dappearance}       &          --                            &     --                                                   & 33.13/.8941                         & 42.22/.9731                                              \\
\multicolumn{1}{|c|}{}                        &
RCAN~\cite{zhang2018image}      & 26.63/.7880                          & 30.07/.8156                                            & {\color[HTML]{3531FF} 36.00/.9123}  & {\color[HTML]{3531FF} 46.33/\color[HTML]{000000}.9791}                       \\
\multicolumn{1}{|c|}{}                        &
MAFFSRN~\cite{muqeet2020ultra}   & 26.57/.7853                          & 30.05/.8141                                            & 35.78/.9101                         & 46.06/.9785                                              \\
\multicolumn{1}{|c|}{}                        &
CSNLN~\cite{mei2020image}     & 26.94/.7958                          &  30.25/.8197                    & 34.24/.9042                         & { 46.11/\color[HTML]{3531FF}.9792}                       \\
\multicolumn{1}{|c|}{}                        &
SRNTT-\textit{$l_2$} & {\color[HTML]{3531FF} 27.62/.8201}   & 30.16/.8176                                            & 35.91/.9120*                         & 46.28/.9791*                                             \\
\multicolumn{1}{|c|}{}                        &
TTSR-\textit{$l_2$}  & 27.02/.8001                          & 29.97/.8123                                            & 35.35/.9083*                         & 37.50/.9709*                                             \\
\multicolumn{1}{|c|}{}                        &
MASA-\textit{$l_2$}  & 27.49/.8145                          & {\color[HTML]{3531FF}30.42/.8263 }                                           & 35.53/.9046*                         &  46.11/.9784*                                           \\
\multicolumn{1}{|c|}{}                        &
AMRSR-\textit{$l_2$}   & {\color[HTML]{FF0000}\textbf{28.32/.8394 }  }                       &{\color[HTML]{FF0000}\textbf{ 30.95/.8438 } }                                         &                         {\color[HTML]{FF0000}\textbf{ 36.82/.9248 }  }          &                                               {\color[HTML]{FF0000}\textbf{46.86/.9814}   }        \\ \hline
\multicolumn{1}{|c|}{\multirow{8}{*}{\rotatebox[origin=c]{90}{\textbf{\textit{Visual-Oriented}}}}}                    & 
SRGAN~\cite{ledig2017photo}     & 23.63/.6761                          & \multicolumn{1}{c}{25.97/.6570}                        & 31.56/.8551                         & \multicolumn{1}{c|}{41.68/.9525}                         \\
\multicolumn{1}{|c|}{}                        &
ESRGAN~\cite{wang2018esrgan}    & 23.69/.6884                          & \multicolumn{1}{c}{26.42/.7005}                        & 32.34/.8664                         & \multicolumn{1}{c|}{42.78/.9669}                         \\
\multicolumn{1}{|c|}{}                        &
RSRGAN~\cite{zhang2019ranksrgan}    & 25.49/.7494                          & \multicolumn{1}{c}{29.10/.7873}                        & 33.90/.8892                         & \multicolumn{1}{c|}{43.44/.9707}                         \\
\multicolumn{1}{|c|}{}                        &
CrossNet~\cite{zheng2018crossnet}  &                 26.00/.7576                     & \multicolumn{1}{c}{29.16/.7834}                                   &                                32.95/.8741     & \multicolumn{1}{c|}{30.30/.9317}                                    \\
\multicolumn{1}{|c|}{}                        &
SSEN~\cite{shim2020robust}  &                 22.71/.7169                     & \multicolumn{1}{c}{26.58/.7824}                                   &                                25.86/.8150     & \multicolumn{1}{c|}{35.61/.9446}                            \\
\multicolumn{1}{|c|}{}                        &
SRNTT~\cite{ZhangSRNTT2019}     & {\color[HTML]{3531FF} 26.42/.7738}   & \multicolumn{1}{c}{{\color[HTML]{3531FF} 29.72/.7984}} & {\color[HTML]{3531FF} 34.78/.8963*}  & \multicolumn{1}{c|}{{\color[HTML]{3531FF} 45.03/.9743*}} \\
\multicolumn{1}{|c|}{}                        &
TTSR~\cite{yang2020learning}      & 25.59/.7645                          & \multicolumn{1}{c}{28.23/.7595}                        & 34.22/.8912*                         & \multicolumn{1}{c|}{35.32/.9566*}                    \\
\multicolumn{1}{|c|}{}                        &
MASA~\cite{lu2021masa}  &                24.84/.7311                   & \multicolumn{1}{c}{27.16/.7129}                                   &                                34.38/.8850*     & \multicolumn{1}{c|}{43.27/.9598*}                                        \\
\multicolumn{1}{|c|}{}                        &
AMRSR       &{\color[HTML]{FF0000} 27.49/.8145  }                        & \multicolumn{1}{c}{{\color[HTML]{FF0000}30.42/.8263} }                       &                \multicolumn{1}{c}{{\color[HTML]{FF0000}35.80/.9122} }                      & \multicolumn{1}{c|}{{\color[HTML]{FF0000}45.56/.9771}}                                \\     \hline
\end{tabular}

%% file: amrsr_arxive/image_files/comparison.tex
\begin{Large}

\begin{tabular}{cccc}
            Ground truth & RCAN~\cite{zhang2018image} & CSNLN~\cite{mei2020image} & RSRGAN~\cite{zhang2019ranksrgan}\\
        \midrule
            Ref. images & TTSR~\cite{yang2020learning} & SRNTT~\cite{ZhangSRNTT2019} & AMRSR (ours)\\
            
             \includegraphics[width=4in]{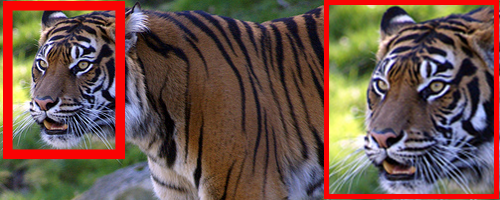} & \includegraphics[width=4in]{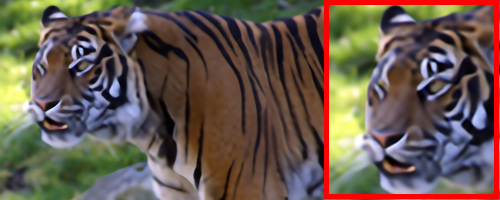} & \includegraphics[width=4in]{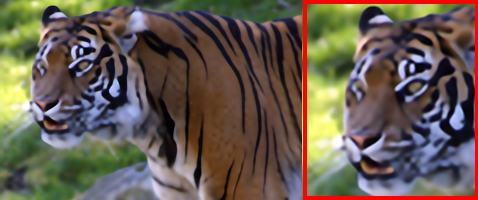} & \includegraphics[width=4in]{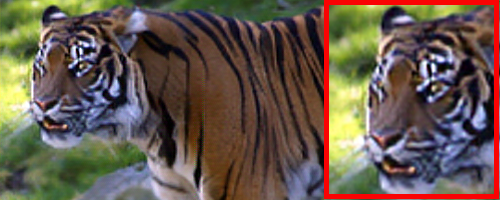} \\
              \includegraphics[width=4in]{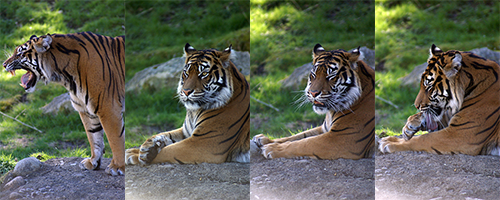} & \includegraphics[width=4in]{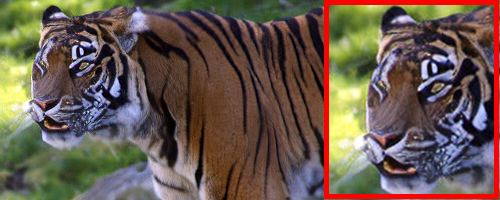} & \includegraphics[width=4in]{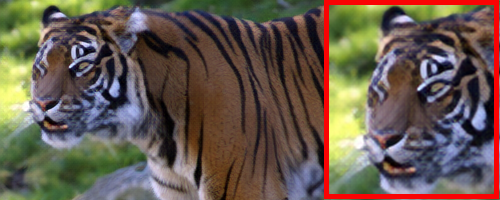} & \includegraphics[width=4in]{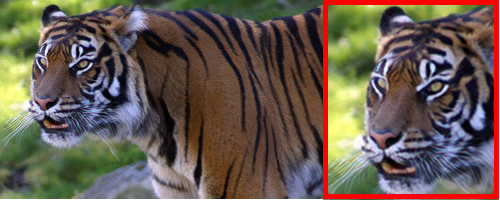} \\

             \includegraphics[width=4in]{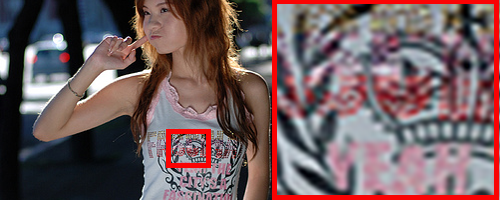} & \includegraphics[width=4in]{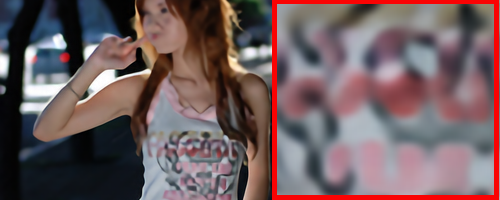} & \includegraphics[width=4in]{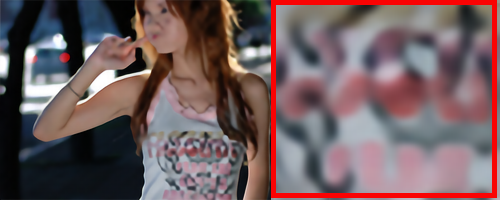} & \includegraphics[width=4in]{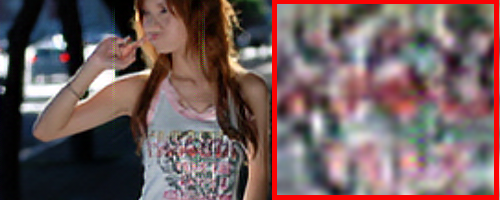} \\
             \includegraphics[width=4in]{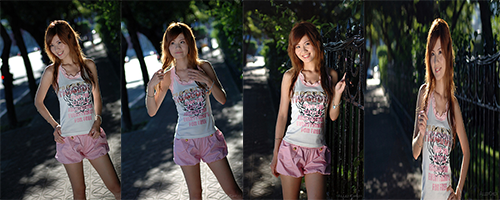} & \includegraphics[width=4in]{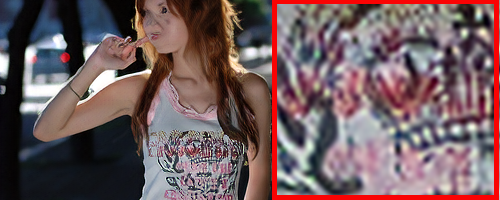} & \includegraphics[width=4in]{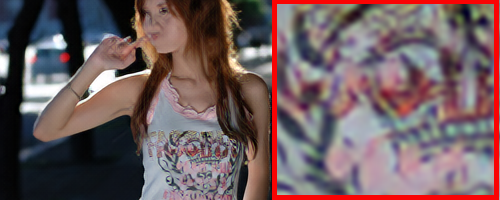} & \includegraphics[width=4in]{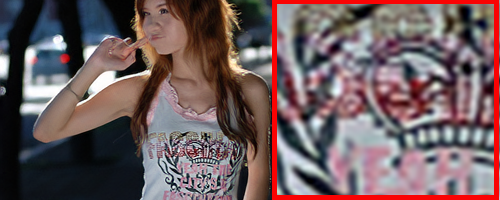} \\
             \midrule 
            \includegraphics[width=4in]{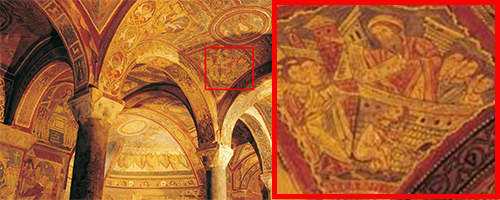} & \includegraphics[width=4in]{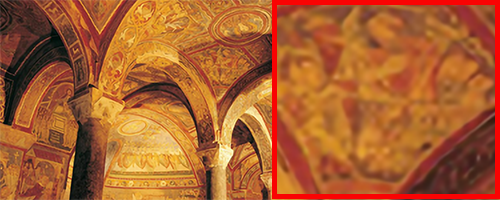} & \includegraphics[width=4in]{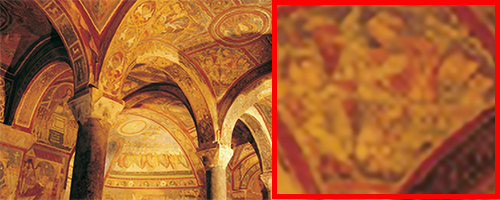} & \includegraphics[width=4in]{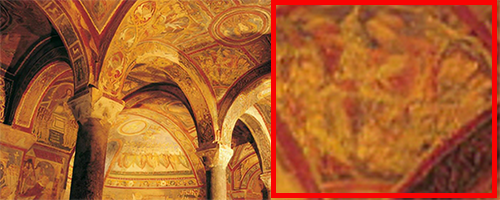} \\
          
              \includegraphics[width=4in]{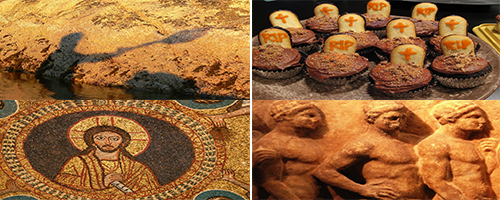} & \includegraphics[width=4in]{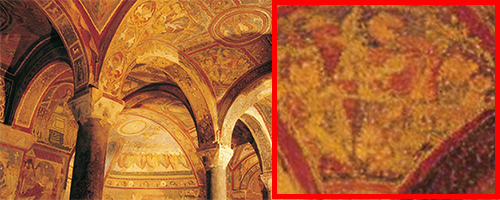} & \includegraphics[width=4in]{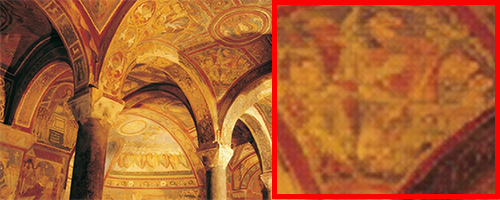} & \includegraphics[width=4in]{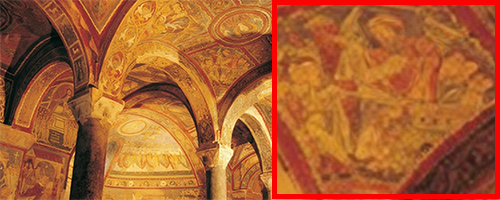} \\
             \includegraphics[width=4in]{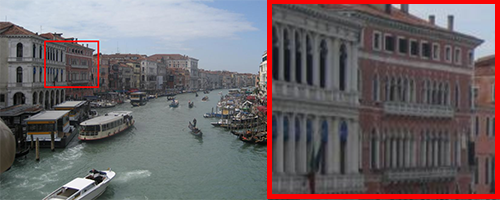} & \includegraphics[width=4in]{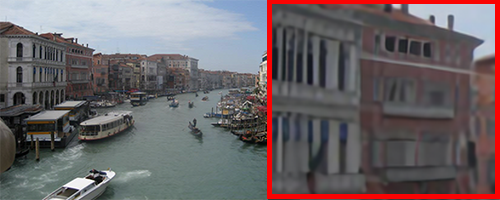} & \includegraphics[width=4in]{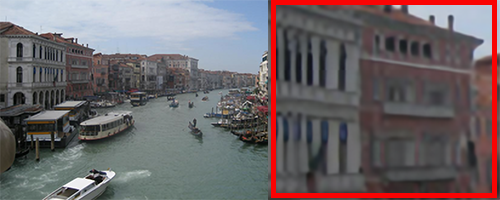} & \includegraphics[width=4in]{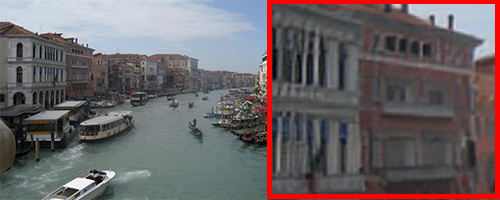} \\
          
             \includegraphics[width=4in]{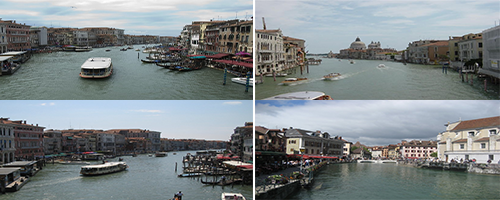} & \includegraphics[width=4in]{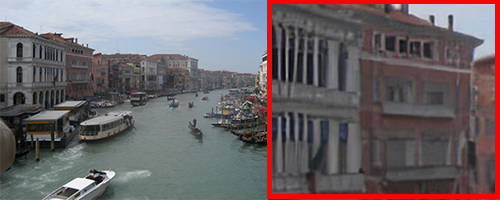} & \includegraphics[width=4in]{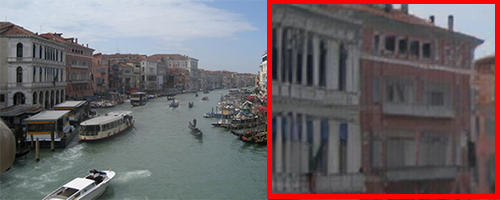} & \includegraphics[width=4in]{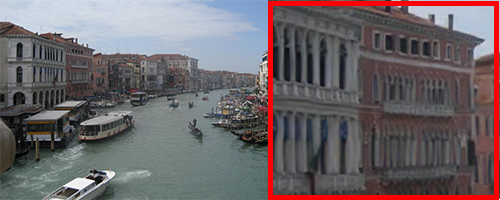} \\
             \midrule
             \includegraphics[width=4in]{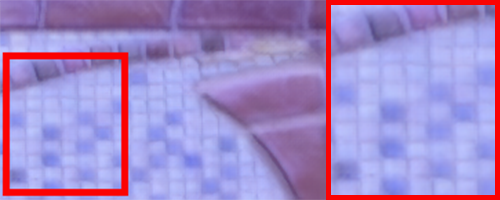} & \includegraphics[width=4in]{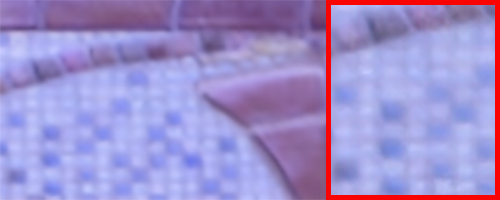} & \includegraphics[width=4in]{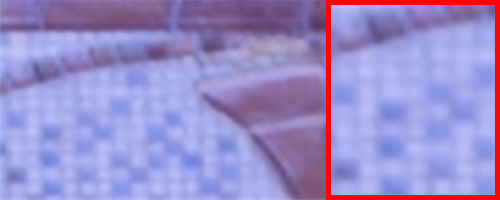} & \includegraphics[width=4in]{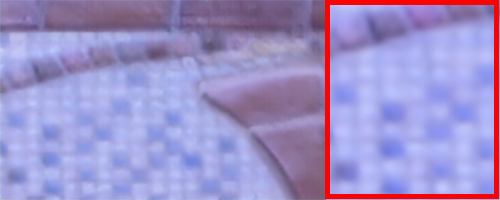} \\
             \includegraphics[width=4in]{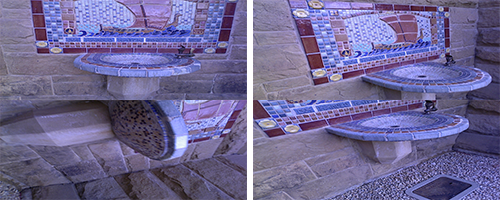} & \includegraphics[width=4in]{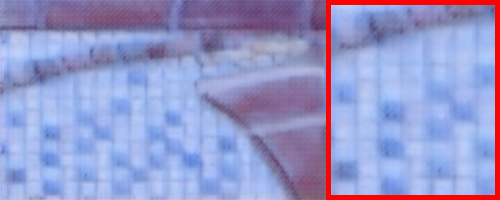} & \includegraphics[width=4in]{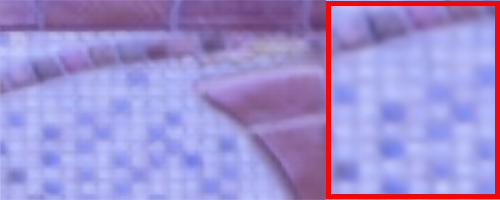} & \includegraphics[width=4in]{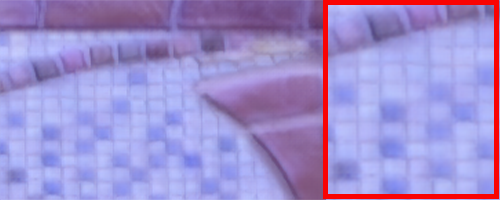} \\
            \midrule
             \includegraphics[width=4in]{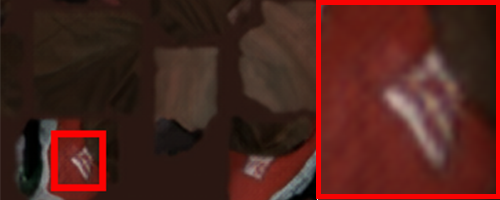} & \includegraphics[width=4in]{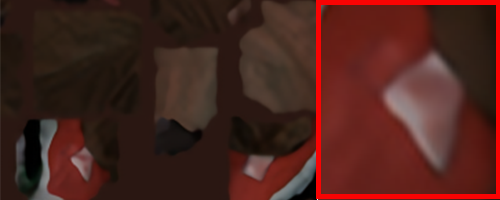} & \includegraphics[width=4in]{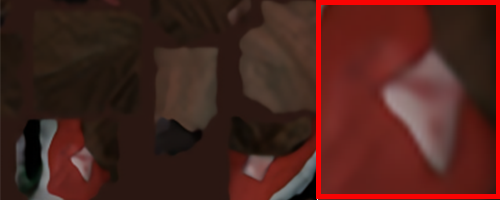} & \includegraphics[width=4in]{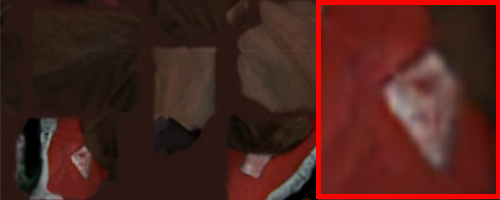} \\
             \includegraphics[width=4in]{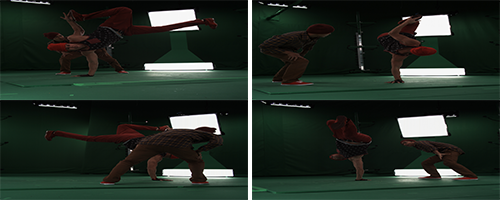} & \includegraphics[width=4in]{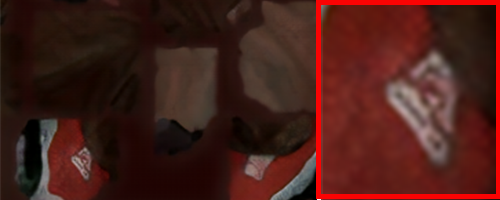} & \includegraphics[width=4in]{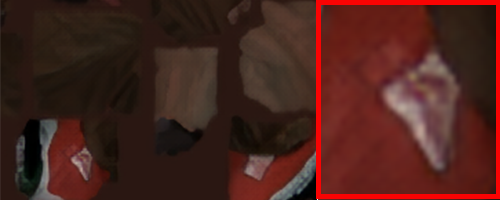} & \includegraphics[width=4in]{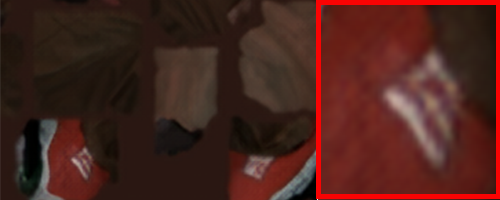} \\
      
        \end{tabular}
\end{Large}

%% file: amrsr_arxive/image_files/chart.tex
\begin{tikzpicture}

    \begin{axis}[
    width=10cm,
    height=6.5cm,
    ybar,
    enlargelimits=0.3,
    bar width=0.9cm,
    ylabel={Percentage (\%)},
    symbolic x coords={
        CSNLN,
        RCAN,
        RSRGAN,
        TTSR,
        SRNTT,
    },
    xtick=data,
    ytick={80,82,84,86,88,90,92,94,96,98,100},
    ]
        \addplot [
            draw=black,
            pattern=vertical lines,
            color=blue,
            grid=major,
            grid style={line width=.1pt, draw=gray!10},
            error bars/.cd,
            y dir=both,
            y explicit,
            error bar style={line width=1pt,color=black}
        ] 
        coordinates
        {
            (CSNLN,97.8)+-(1.16,1.16)
            (RCAN,98.5)+-(0.97,0.97)
            (RSRGAN,98.1)+-(1.07,1.07)
            (TTSR,90.1)+-(2.38,2.38)
            (SRNTT,97.1)+-(1.32,1.32)
        };
        
        \node at (axis description cs: 0.18,.83) {\scriptsize 97.8$\pm$1.16};
        \node at (axis description cs: 0.35,.87) {\scriptsize 98.5$\pm$0.97};
        \node at (axis description cs: 0.5,.83) {\scriptsize 98.1$\pm$1.07};
        \node at (axis description cs: 0.65,.48) {\scriptsize 90.1$\pm$2.38};
        \node at (axis description cs: 0.82,.79) {\scriptsize 97.1$\pm$1.32};
    
    \end{axis}
\end{tikzpicture}

%% file: amrsr_arxive/table/simlevel_v2.tex
\begin{tabular}{|c|cccc|}
\hline
\multicolumn{1}{|c|}{\bf Algorithms} & \multicolumn{1}{c|}{\bf L1} & \multicolumn{1}{c|}{\bf L2} & \multicolumn{1}{c|}{\bf L3} & \bf L4                   \\ \hline
Cross-Net~\cite{zheng2018crossnet}                        & 25.98/.7582             & 25.98/.7582             & 25.97/.7581             & 25.97/.7581          \\
SSEN~\cite{shim2020robust}                     & 22.71/.7169             & 22.43/.7114             & 22.30/.7131             & 22.13/.7084          \\
SRNTT~\cite{ZhangSRNTT2019}                           & 26.42/.7738             & 26.34/.7690             & 26.27/.7682             & 26.24/.7678          \\
TTSR~\cite{yang2020learning}                             & 25.59/.7645             & 25.08/.7442             & 24.98/.7414             & 24.95/.7412        \\
MASA~\cite{lu2021masa}                             & 24.84/.7311             & 24.27/.7093            & 24.25/.7077             & 24.23/.7057         \\
AMRSR 1ref                         & 26.77/.7882             & 26.71/.7869             & 26.63/.7841             & 26.48/.7804          \\ \hline
SRNTT-\textit{$l_2$}                       & 27.62/.8201             & 27.21/.8039             & 27.05/.8003             & 26.92/.7969          \\
TTSR-\textit{$l_2$}                         & 27.02/.8001             & 26.48/.7809             & 26.40/.7792             & 26.35/.7784         \\
MASA-\textit{$l_2$}                         &     27.49/.8145         &      26.66/.7881        & 26.60/.7863             & 26.55/.7843         \\
AMRSR 1-\textit{$l_2$}                        & \textbf{27.92/.8293}    & \textbf{27.51/.8152}    & \textbf{27.40/.8114}    & \textbf{27.24/.8080} \\ \hline
\end{tabular}

%% file: amrsr_arxive/table/sim2_v2.tex
\begin{tabular}{|c|ccccccc|}
\hline
       & \multicolumn{1}{c|}{TTSR~\cite{yang2020learning}} & \multicolumn{1}{c|}{SRNTT~\cite{ZhangSRNTT2019}} & \multicolumn{1}{c|}{CrossNet~\cite{zheng2018crossnet}} & \multicolumn{1}{c|}{SSEN~\cite{shim2020robust}}& \multicolumn{1}{c|}{MASA~\cite{lu2021masa}} & \multicolumn{1}{c|}{AMRSR 1} & AMRSR         \\ \hline
Visual & 25.08/.746               & 26.17/.765              & 25.98/.758                   & 22.83/.7148               &24.17/.703& 26.36/.775                              & \textbf{26.47/.780} \\
PSNR   & 26.59/.784               & 26.67/.790          & --                            & --                        &26.42/.780& 27.06/.805                               & \textbf{27.10/.806} \\ \hline
\end{tabular}

%% file: amrsr_arxive/image_files/sim.tex
\begin{tabular}{cccc}
\shortstack{Ground truth\\Ref. images}   & \shortstack{SRNTT\\AMRSR 1} & \shortstack{TTSR \\AMRSR}\\

\shortstack{\includegraphics[width=2.2in]{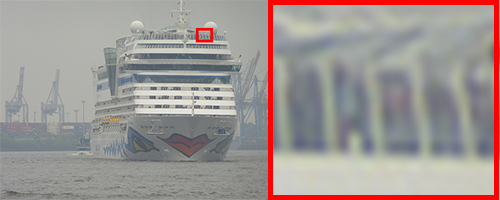} \\\includegraphics[width=2.2in]{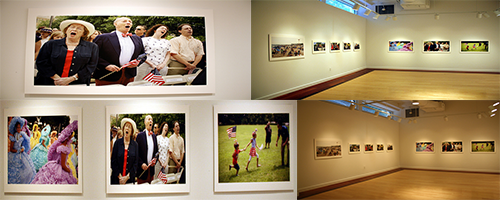} }      &   \shortstack{\includegraphics[width=1in]{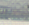}  \\ \includegraphics[width=1in]{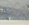}  }  &   \shortstack{ \includegraphics[width=1in]{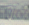} \\  \includegraphics[width=1in]{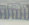}}  
\end{tabular}

%% file: amrsr_arxive/table/cimr2.tex
\begin{tabular}{|c|cl|cl|}
\hline
\multirow{2}{*}{\textbf{Algorithms}} & \multicolumn{2}{c|}{\textbf{Visual-oriented}}                 & \multicolumn{2}{c|}{\textbf{PSNR-oriented}}                   \\ \cline{2-5} 
                                     & \multicolumn{1}{c|}{CUFED5} & \multicolumn{1}{c|}{Sun80}      & \multicolumn{1}{c|}{CUFED5} & \multicolumn{1}{c|}{Sun80}      \\ \hline
SRNTT~\cite{ZhangSRNTT2019}                              & 25.61/.764                  & \multicolumn{1}{c|}{27.59/.756} & 26.24/.784                  & \multicolumn{1}{c|}{28.54/.793} \\
CIMR~\cite{cui2020towards}                                 & 26.16/.781                  & 29.67/.806                      & 26.35/.789                  & 30.07/.813                     \\
AMRSR $N_M=4$                          & 26.87/.795                  & 30.53/.829                      & 27.57/.820                  & 31.10/.847                      \\
AMRSR $N_M=8$                          & 26.92/.796                  & 30.64/.832                      & 27.63/.821                  & 31.24/.851                      \\
AMRSR $N_M=16$                         & 26.98/.797                  & 30.69/.834                      & 27.69/.823                  & 31.29/.852                      \\
AMRSR $N_M=32$                         & 27.01/.798                  &         30.75/.835                        & 27.75/.824                  &         31.35/.854                        \\
AMRSR $N_M=64$                         & \textbf{27.08/.800}         &         \textbf{30.80/.836}                        & \textbf{27.81/.826}         &                \textbf{31.41/.854}                 \\ \hline
\end{tabular}

%% file: amrsr_arxive/table/reference_psnr2.tex
\begin{tabular}{c|c|cccc|}
\cline{2-6}
&
\textbf{Nr. references} & \textbf{CU4REF} & \textbf{Sun80} & \textbf{GEMAP} & \textbf{HUMAP} \\ \hline
\multicolumn{1}{|c|}{\multirow{4}{*}{\rotatebox[origin=c]{90}{\textbf{\textit{\small PSNR-Oriented}}}}} &
1 Reference                                          & 27.92/.8293                & 30.61/.8376               &              36.49/.9219             &                    46.64/.9803        \\
\multicolumn{1}{|c|}{}  &
2 References                                          & 28.26/.8384                & 30.79/.8380               & 36.58/.9230              & 46.74/.9808                \\
\multicolumn{1}{|c|}{}  &
4 References                               & \textbf{28.32/.8394 }               & \textbf{30.95/.8438}             &         36.82/.9248               & 46.86/.9814                \\
\multicolumn{1}{|c|}{}  &
8 References                                             & --                         & --                        & \textbf{36.84/.9255 }              &             \textbf{46.87/.9814 }              \\

\cline{2-6}
\multicolumn{1}{|c|}{}  &
$2^{nd}$ best                                 & 27.62/.8201                & 30.25/.8197               & 36.00/.9123               & 46.33/.9792                \\ \hline
\end{tabular}

%% file: amrsr_arxive/table/attention_psnr2.tex
\begin{tabular}{c|c|cccc|}
\cline{2-6}
&
\textbf{Config.} &\textbf{CU4REF} & \textbf{Sun80} & \textbf{GEMAP} & \textbf{HUMAP} \\ \hline
\multicolumn{1}{|c|}{\multirow{4}{*}{\rotatebox[origin=c]{90}{\textbf{\textit{\scriptsize	 PSNR-Oriented}}}}} &
No attention                                     & 27.54/.8170                & 30.18/.8171            & 35.91/.9130               & 46.33/.9791                \\
\multicolumn{1}{|c|}{}  &
Ref. attention                                  & 27.55/.8179                &             30.21/.8196              & 35.93/.9131               & 46.43/.9794                \\
\multicolumn{1}{|c|}{}  &
Both attention                                    & 27.18/.8064                & 30.24/.8194               & 35.91/.9130               & 46.43/.9793                \\
\multicolumn{1}{|c|}{}  &
AMRSR                                 & \textbf{28.32/.8394 }               & \textbf{30.95/.8438   }            &          \textbf{36.82/.9248}                 & \textbf{46.86/.9814}       \\ \hline
\end{tabular}

%% file: amrsr_arxive/image_files/attention.tex
\begin{tabular}{cccc}
\shortstack{Ground truth\\Ref. images}   & \shortstack{No att.\\Both att.} & \shortstack{Ref. att. \\AMRSR}\\

\shortstack{\includegraphics[width=2.5in]{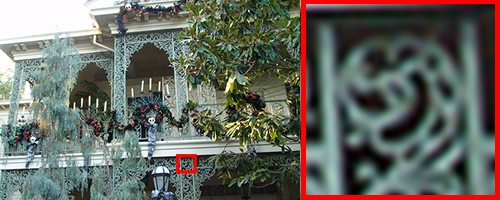} \\\includegraphics[width=2.5in]{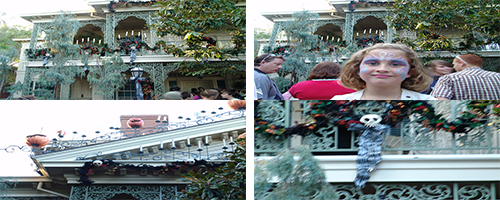} }      &   \shortstack{\includegraphics[width=1in]{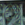}  \\ \includegraphics[width=1in]{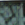}  }  &   \shortstack{ \includegraphics[width=1in]{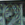} \\  \includegraphics[width=1in]{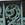}}  
\end{tabular}

%% file: amrsr_arxive/image_files/reference2.tex
%
\begin{tabular}{ccc}
\shortstack{Ground truth\\Ref. images}   & \shortstack{1 ref.\\4 ref.} & \shortstack{2 ref. \\8 ref.}\\
                              
   \shortstack{\includegraphics[width=3in]{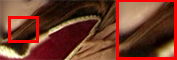} \\\includegraphics[width=3in]{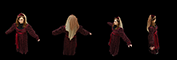} }      &   \shortstack{\includegraphics[width=1in]{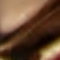}  \\ \includegraphics[width=1in]{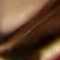}  }  &   \shortstack{ \includegraphics[width=1in]{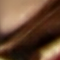} \\  \includegraphics[width=1in]{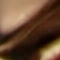}}  \\

    \shortstack{\includegraphics[width=3in]{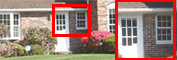} \\ \includegraphics[width=3in]{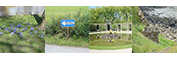}}    &   \shortstack{\includegraphics[width=1in]{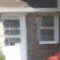}\\  \includegraphics[width=1in]{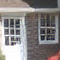}}  &    \shortstack{\includegraphics[width=1in]{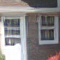} \\  \includegraphics[width=1in]{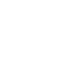}} \\
                             
\end{tabular}

%% file: amrsr_arxive/table/memory3_v2.tex
\begin{tabular}{|c|cc|cc|cc|cc|}
\hline
\multirow{2}{*}{\textbf{Algorithms}} & \multicolumn{2}{c|}{\textbf{CU4REF}}                                    & \multicolumn{2}{c|}{\textbf{Sun80}}                                     & \multicolumn{2}{c|}{\textbf{GEMAP}}                                     & \multicolumn{2}{c|}{\textbf{HUMAP}}                                     \\ \cline{2-9} 
                                  & \footnotesize{PSNR/SSIM} & \footnotesize{GPU} &  \footnotesize{PSNR/SSIM} & \footnotesize{GPU} &  \footnotesize{PSNR/SSIM} & \footnotesize{GPU} & \footnotesize{PSNR/SSIM} & \footnotesize{GPU}  \\ \hline
AMRSR 1                          & 27.92/.829                   & \textbf{1.01}                          & 30.61/.837                 & \textbf{2.10}                         &              36.49/.921           & 15.57                          &        46.64/.980               & 11.47                          \\
AMRSR 2                          & 28.26/.838                   & 1.25                       & 30.79/.838            & 2.66                          & 36.56/.922           & 15.63                     & 46.74/.980                 & 11.53                      \\
AMRSR 8                         & --                             & --                                 & --                             & --                              & \textbf{36.84/.925}                    & 15.80                        &               \textbf{46.87/.981}               & 11.70                    \\
$N_{R}=1$                        & \textbf{28.32/.839}                    & 1.36                         & 30.84/.839               & 3.96                         &            36.60/.922                & 40.15                    &              46.81/.981                 & 29.60                        \\
$N_{R}=4$                        & 28.00/.835                    & 1.21                        & 30.87/.841                    & 3.23                      & 36.65/.923                   & 28.49                     & 46.85/.981                    & 20.98                      \\
$N_{R}=16$                       & 27.97/.834                    & 1.22                        & \textbf{30.95/.843}                    & 3.24                        &             36.82/.924                   & 15.69              & 46.86/.981                    & 11.59                         \\
cut 1                      & --                             & --                               & --                             & --                              &              36.44/.920              & \textbf{3.99}                       &                46.63/.980                & \textbf{2.94}              \\
cut 4                      & --                             & --                             & --                             & --                              &                 36.75/.924               & 4.10                         & 46.81/.981                    & 3.06                       \\
SRNTT~\cite{ZhangSRNTT2019}                             & 27.62/.820                    & 2.81                           & 30.16/.817                   & 14.63                         & 35.91/.912                    & (26.51)              & 46.28/.979                    & (19.49)                \\
TTSR~\cite{yang2020learning}                              & 27.02/.800                    & 4.24                        & 29.97/.812                    & 20.61                    & 35.35/.908                    &  (40.07)                 & 37.50/.970                   & (29.47)               \\

MASA~\cite{lu2021masa}                              & 27.49/.814                    & 8.23                        & 30.42/.826                    & 15.69                    & 35.53/.904                    &  (42.11)                 & 46.11/.978                   & (31.49)               \\
\hline
\end{tabular}

%% file: camera_ready_v2.bbl
\begin{thebibliography}{10}\itemsep=-1pt

\bibitem{blender}
Blender.
\newblock https://www.blender.org/.
\newblock Accessed: 2020-12-26.

\bibitem{cgtrader}
Cgtrader.
\newblock hhttps://www.cgtrader.com/free-3d-models.
\newblock Accessed: 2020-12-26.

\bibitem{free3d}
Free3d.
\newblock https://free3d.com/3d-models/.
\newblock Accessed: 2020-12-26.

\bibitem{renderpeople}
Renderpeople.
\newblock https://renderpeople.com/.
\newblock Accessed: 2020-12-26.

\bibitem{turbosquid}
Turbosquid.
\newblock https://www.turbosquid.com/Search/3D-Models/free.
\newblock Accessed: 2020-12-26.

\bibitem{boominathan2014improving}
Vivek Boominathan, Kaushik Mitra, and Ashok Veeraraghavan.
\newblock Improving resolution and depth-of-field of light field cameras using
  a hybrid imaging system.
\newblock In {\em 2014 IEEE International Conference on Computational
  Photography (ICCP)}, pages 1--10. IEEE, 2014.

\bibitem{collet2015high}
Alvaro Collet, Ming Chuang, Pat Sweeney, Don Gillett, Dennis Evseev, David
  Calabrese, Hugues Hoppe, Adam Kirk, and Steve Sullivan.
\newblock High-quality streamable free-viewpoint video.
\newblock {\em ACM Transactions on Graphics (ToG)}, 34(4):1--13, 2015.

\bibitem{cui2020towards}
Shuguang Cui.
\newblock Towards content-independent multi-reference super-resolution:
  Adaptive pattern matching and feature aggregation.
\newblock 2020.

\bibitem{dong2014learning}
Chao Dong, Chen~Change Loy, Kaiming He, and Xiaoou Tang.
\newblock Learning a deep convolutional network for image super-resolution.
\newblock In {\em European conference on computer vision}, pages 184--199.
  Springer, 2014.

\bibitem{goldlucke2014super}
Bastian Goldl{\"u}cke, Mathieu Aubry, Kalin Kolev, and Daniel Cremers.
\newblock A super-resolution framework for high-accuracy multiview
  reconstruction.
\newblock {\em International journal of computer vision}, 106(2):172--191,
  2014.

\bibitem{gulrajani2017improved}
Ishaan Gulrajani, Faruk Ahmed, Martin Arjovsky, Vincent Dumoulin, and Aaron~C
  Courville.
\newblock Improved training of wasserstein gans.
\newblock In {\em Advances in neural information processing systems}, pages
  5767--5777, 2017.

\bibitem{he2016deep}
Kaiming He, Xiangyu Zhang, Shaoqing Ren, and Jian Sun.
\newblock Deep residual learning for image recognition.
\newblock In {\em Proceedings of the IEEE conference on computer vision and
  pattern recognition}, pages 770--778, 2016.

\bibitem{johnson2016perceptual}
Justin Johnson, Alexandre Alahi, and Li Fei-Fei.
\newblock Perceptual losses for real-time style transfer and super-resolution.
\newblock In {\em European conference on computer vision}, pages 694--711.
  Springer, 2016.

\bibitem{kim2016accurate}
Jiwon Kim, Jung Kwon~Lee, and Kyoung Mu~Lee.
\newblock Accurate image super-resolution using very deep convolutional
  networks.
\newblock In {\em Proceedings of the IEEE conference on computer vision and
  pattern recognition}, pages 1646--1654, 2016.

\bibitem{ledig2017photo}
Christian Ledig, Lucas Theis, Ferenc Husz{\'a}r, Jose Caballero, Andrew
  Cunningham, Alejandro Acosta, Andrew Aitken, Alykhan Tejani, Johannes Totz,
  Zehan Wang, et~al.
\newblock Photo-realistic single image super-resolution using a generative
  adversarial network.
\newblock In {\em Proceedings of the IEEE conference on computer vision and
  pattern recognition}, pages 4681--4690, 2017.

\bibitem{ley2016syb3r}
Andreas Ley, Ronny H{\"a}nsch, and Olaf Hellwich.
\newblock Syb3r: A realistic synthetic benchmark for 3d reconstruction from
  images.
\newblock In {\em European Conference on Computer Vision}, pages 236--251.
  Springer, 2016.

\bibitem{li2020survey}
Kai Li, Shenghao Yang, Runting Dong, Xiaoying Wang, and Jianqiang Huang.
\newblock Survey of single image super-resolution reconstruction.
\newblock {\em IET Image Processing}, 14(11):2273--2290, 2020.

\bibitem{li2019_3dappearance}
Yawei Li, Vagia Tsiminaki, Radu Timofte, Marc Pollefeys, and Luc Van~Gool.
\newblock 3d appearance super-resolution with deep learning.
\newblock In {\em In Proceedings of the IEEE International Conference on
  Computer Vision}, 2019.

\bibitem{lim2017enhanced}
Bee Lim, Sanghyun Son, Heewon Kim, Seungjun Nah, and Kyoung Mu~Lee.
\newblock Enhanced deep residual networks for single image super-resolution.
\newblock In {\em Proceedings of the IEEE conference on computer vision and
  pattern recognition workshops}, pages 136--144, 2017.

\bibitem{liu2020residual}
Jie Liu, Jie Tang, and Gangshan Wu.
\newblock Residual feature distillation network for lightweight image
  super-resolution.
\newblock {\em arXiv preprint arXiv:2009.11551}, 2020.

\bibitem{lu2021masa}
Liying Lu, Wenbo Li, Xin Tao, Jiangbo Lu, and Jiaya Jia.
\newblock Masa-sr: Matching acceleration and spatial adaptation for
  reference-based image super-resolution.
\newblock {\em arXiv preprint arXiv:2106.02299}, 2021.

\bibitem{mei2020image}
Yiqun Mei, Yuchen Fan, Yuqian Zhou, Lichao Huang, Thomas~S Huang, and Honghui
  Shi.
\newblock Image super-resolution with cross-scale non-local attention and
  exhaustive self-exemplars mining.
\newblock In {\em Proceedings of the IEEE/CVF Conference on Computer Vision and
  Pattern Recognition}, pages 5690--5699, 2020.

\bibitem{muqeet2020ultra}
Abdul Muqeet, Jiwon Hwang, Subin Yang, Jung~Heum Kang, Yongwoo Kim, and Sung-Ho
  Bae.
\newblock Ultra lightweight image super-resolution with multi-attention layers.
\newblock {\em arXiv preprint arXiv:2008.12912}, 2020.

\bibitem{pandey2018compendious}
Garima Pandey and Umesh Ghanekar.
\newblock A compendious study of super-resolution techniques by single image.
\newblock {\em Optik}, 166:147--160, 2018.

\bibitem{richard2019learned}
Audrey Richard, Ian Cherabier, Martin~R Oswald, Vagia Tsiminaki, Marc
  Pollefeys, and Konrad Schindler.
\newblock Learned multi-view texture super-resolution.
\newblock In {\em 2019 International Conference on 3D Vision (3DV)}, pages
  533--543. IEEE, 2019.

\bibitem{schops2017multi}
Thomas Schops, Johannes~L Schonberger, Silvano Galliani, Torsten Sattler,
  Konrad Schindler, Marc Pollefeys, and Andreas Geiger.
\newblock A multi-view stereo benchmark with high-resolution images and
  multi-camera videos.
\newblock In {\em Proceedings of the IEEE Conference on Computer Vision and
  Pattern Recognition}, pages 3260--3269, 2017.

\bibitem{seitz2006comparison}
Steven~M Seitz, Brian Curless, James Diebel, Daniel Scharstein, and Richard
  Szeliski.
\newblock A comparison and evaluation of multi-view stereo reconstruction
  algorithms.
\newblock In {\em 2006 IEEE computer society conference on computer vision and
  pattern recognition (CVPR'06)}, volume~1, pages 519--528. IEEE, 2006.

\bibitem{shim2020robust}
Gyumin Shim, Jinsun Park, and In~So Kweon.
\newblock Robust reference-based super-resolution with similarity-aware
  deformable convolution.
\newblock In {\em Proceedings of the IEEE/CVF Conference on Computer Vision and
  Pattern Recognition}, pages 8425--8434, 2020.

\bibitem{simonyan2014very}
Karen Simonyan and Andrew Zisserman.
\newblock Very deep convolutional networks for large-scale image recognition.
\newblock {\em arXiv preprint arXiv:1409.1556}, 2014.

\bibitem{sun2012super}
Libin Sun and James Hays.
\newblock Super-resolution from internet-scale scene matching.
\newblock In {\em 2012 IEEE International Conference on Computational
  Photography (ICCP)}, pages 1--12. IEEE, 2012.

\bibitem{tai2017image}
Ying Tai, Jian Yang, and Xiaoming Liu.
\newblock Image super-resolution via deep recursive residual network.
\newblock In {\em Proceedings of the IEEE conference on computer vision and
  pattern recognition}, pages 3147--3155, 2017.

\bibitem{tan2020crossnet++}
Yang Tan, Haitian Zheng, Yinheng Zhu, Xiaoyun Yuan, Xing Lin, David Brady, and
  Lu Fang.
\newblock Crossnet++: Cross-scale large-parallax warping for reference-based
  super-resolution.
\newblock {\em IEEE Computer Architecture Letters}, (01):1--1, 2020.

\bibitem{tong2017image}
Tong Tong, Gen Li, Xiejie Liu, and Qinquan Gao.
\newblock Image super-resolution using dense skip connections.
\newblock In {\em Proceedings of the IEEE International Conference on Computer
  Vision}, pages 4799--4807, 2017.

\bibitem{wang2018recovering}
Xintao Wang, Ke Yu, Chao Dong, and Chen~Change Loy.
\newblock Recovering realistic texture in image super-resolution by deep
  spatial feature transform.
\newblock In {\em Proceedings of the IEEE conference on computer vision and
  pattern recognition}, pages 606--615, 2018.

\bibitem{wang2018esrgan}
Xintao Wang, Ke Yu, Shixiang Wu, Jinjin Gu, Yihao Liu, Chao Dong, Yu Qiao, and
  Chen Change~Loy.
\newblock Esrgan: Enhanced super-resolution generative adversarial networks.
\newblock In {\em Proceedings of the European Conference on Computer Vision
  (ECCV)}, pages 0--0, 2018.

\bibitem{wang2016event}
Yufei Wang, Zhe Lin, Xiaohui Shen, Radomir Mech, Gavin Miller, and Garrison~W
  Cottrell.
\newblock Event-specific image importance.
\newblock In {\em Proceedings of the IEEE Conference on Computer Vision and
  Pattern Recognition}, pages 4810--4819, 2016.

\bibitem{wang2020deep}
Zhihao Wang, Jian Chen, and Steven~CH Hoi.
\newblock Deep learning for image super-resolution: A survey.
\newblock {\em IEEE Transactions on Pattern Analysis and Machine Intelligence},
  2020.

\bibitem{yang2014single}
Chih-Yuan Yang, Chao Ma, and Ming-Hsuan Yang.
\newblock Single-image super-resolution: A benchmark.
\newblock In {\em European Conference on Computer Vision}, pages 372--386.
  Springer, 2014.

\bibitem{yang2020learning}
Fuzhi Yang, Huan Yang, Jianlong Fu, Hongtao Lu, and Baining Guo.
\newblock Learning texture transformer network for image super-resolution.
\newblock In {\em Proceedings of the IEEE/CVF Conference on Computer Vision and
  Pattern Recognition}, pages 5791--5800, 2020.

\bibitem{yu2018wide}
Jiahui Yu, Yuchen Fan, Jianchao Yang, Ning Xu, Zhaowen Wang, Xinchao Wang, and
  Thomas Huang.
\newblock Wide activation for efficient and accurate image super-resolution.
\newblock {\em arXiv preprint arXiv:1808.08718}, 2018.

\bibitem{yue2013landmark}
Huanjing Yue, Xiaoyan Sun, Jingyu Yang, and Feng Wu.
\newblock Landmark image super-resolution by retrieving web images.
\newblock {\em IEEE Transactions on Image Processing}, 22(12):4865--4878, 2013.

\bibitem{zhang2019ranksrgan}
Wenlong Zhang, Yihao Liu, Chao Dong, and Yu Qiao.
\newblock Ranksrgan: Generative adversarial networks with ranker for image
  super-resolution.
\newblock In {\em Proceedings of the IEEE International Conference on Computer
  Vision}, pages 3096--3105, 2019.

\bibitem{zhang2018image}
Yulun Zhang, Kunpeng Li, Kai Li, Lichen Wang, Bineng Zhong, and Yun Fu.
\newblock Image super-resolution using very deep residual channel attention
  networks.
\newblock In {\em Proceedings of the European Conference on Computer Vision
  (ECCV)}, pages 286--301, 2018.

\bibitem{ZhangSRNTT2019}
Zhifei Zhang, Zhaowen Wang, Zhe Lin, and Hairong Qi.
\newblock Image super-resolution by neural texture transfer.
\newblock In {\em Proceedings of the IEEE Conference on Computer Vision and
  Pattern Recognition}, pages 7982--7991, 2019.

\bibitem{zheng2017learning}
Haitian Zheng, Mengqi Ji, Lei Han, Ziwei Xu, Haoqian Wang, Yebin Liu, and Lu
  Fang.
\newblock Learning cross-scale correspondence and patch-based synthesis for
  reference-based super-resolution.
\newblock In {\em BMVC}, 2017.

\bibitem{zheng2018crossnet}
Haitian Zheng, Mengqi Ji, Haoqian Wang, Yebin Liu, and Lu Fang.
\newblock Crossnet: An end-to-end reference-based super resolution network
  using cross-scale warping.
\newblock In {\em Proceedings of the European Conference on Computer Vision
  (ECCV)}, pages 88--104, 2018.

\bibitem{zhou2014color}
Qian-Yi Zhou and Vladlen Koltun.
\newblock Color map optimization for 3d reconstruction with consumer depth
  cameras.
\newblock {\em ACM Transactions on Graphics (TOG)}, 33(4):1--10, 2014.

\bibitem{zollhofer2015shading}
Michael Zollh{\"o}fer, Angela Dai, Matthias Innmann, Chenglei Wu, Marc
  Stamminger, Christian Theobalt, and Matthias Nie{\ss}ner.
\newblock Shading-based refinement on volumetric signed distance functions.
\newblock {\em ACM Transactions on Graphics (TOG)}, 34(4):1--14, 2015.

\end{thebibliography}
